\documentclass{article}

\usepackage{booktabs}   
\usepackage{siunitx}    
\usepackage{xcolor}     
\usepackage{colortbl}   
\usepackage{multirow}   

\usepackage{hyperref}
\hypersetup{
    colorlinks=true,
    linkcolor=black,
    filecolor=magenta,      
    urlcolor=blue,
    citecolor=black,
}

\usepackage{pifont}
\usepackage{enumitem} 
\usepackage{tabularx}
\usepackage{booktabs} 

\usepackage{hyperref}
\usepackage{url}
\usepackage{subfigure}
\usepackage{wrapfig}
\usepackage{microtype}
\usepackage{graphicx}
\usepackage{subcaption}
\usepackage{xcolor}
\usepackage{amsmath} 
\definecolor{PhysicsBlue}{RGB}{0, 82, 156}      
\definecolor{NeuralOchre}{RGB}{204, 136, 0}     

\definecolor{PhysicsTeal}{RGB}{0, 128, 128}     
\definecolor{NeuralPurple}{RGB}{106, 76, 147}   
\usepackage{hyperref}

\usepackage[preprint]{icml2026}

\usepackage{amsmath}
\usepackage{amssymb}
\usepackage{mathtools}
\usepackage{amsthm}

\usepackage[capitalize,noabbrev]{cleveref}

\theoremstyle{plain}

\theoremstyle{definition}

\theoremstyle{remark}

\usepackage[textsize=tiny]{todonotes}

\icmltitlerunning{NeuralOGCM}

\begin{document}
\newcommand{\method}{{\fontfamily{lmtt}\selectfont \textbf{NeuralOGCM}}}

\twocolumn[
  \icmltitle{\method{}: Differentiable Ocean Modeling with Learnable Physics}

  \icmlsetsymbol{equal}{$\dagger$}

  \begin{icmlauthorlist}
    \icmlauthor{Hao Wu}{thu,equal}
    \icmlauthor{Yuan Gao}{thu,equal}
    \icmlauthor{Fan Xu}{ustc,equal}
    \icmlauthor{Fan Zhang}{cuhk}
    \icmlauthor{Guangliang Liu}{thu}
    \icmlauthor{Yuxuan Liang}{hkust}
    \icmlauthor{Xiaomeng Huang}{thu}
  \end{icmlauthorlist}

  \icmlaffiliation{thu}{Tsinghua University, Beijing, China}
  \icmlaffiliation{ustc}{University of Science and Technology of China, Hefei, China}
  \icmlaffiliation{cuhk}{The Chinese University of Hong Kong, Hong Kong, China}
\icmlaffiliation{hkust}{Hong Kong University of Science and Technology (Guangzhou), Guangzhou, China}
  \icmlcorrespondingauthor{Xiaomeng Huang}{hxm@tsinghua.edu.cn}

  \icmlkeywords{Machine Learning, ICML}

  \vskip 0.3in
]


\printAffiliationsAndNotice{}  

\begin{abstract}
High-precision scientific simulation faces a long-standing trade-off between computational efficiency and physical fidelity. 
To address this challenge, we propose \method{}, an ocean modeling framework that fuses differentiable programming with deep learning. 
At the core of \method{} is a fully differentiable dynamical solver, which leverages physics knowledge as its core inductive bias. 
The learnable physics integration captures large-scale, deterministic physical evolution, and transforms key physical parameters (e.g., diffusion coefficients) into learnable parameters, enabling the model to autonomously optimize its physical core via end-to-end training. 
Concurrently, a deep neural network learns to correct for subgrid-scale processes and discretization errors not captured by the physics model. 
Both components work in synergy, with their outputs integrated by a unified ODE solver. 
Experiments demonstrate that \method{} maintains long-term stability and physical consistency, significantly outperforming traditional numerical models in speed and pure AI baselines in accuracy. 
Our work paves a new path for building fast, stable, and physically-plausible models for scientific computing. Our codes are available \href{https://github.com/Alexander-wu/NeuralOGCM}{here}.
\end{abstract}
\section{Introduction}
\begin{figure}
  \centering
  \includegraphics[width=1\linewidth]{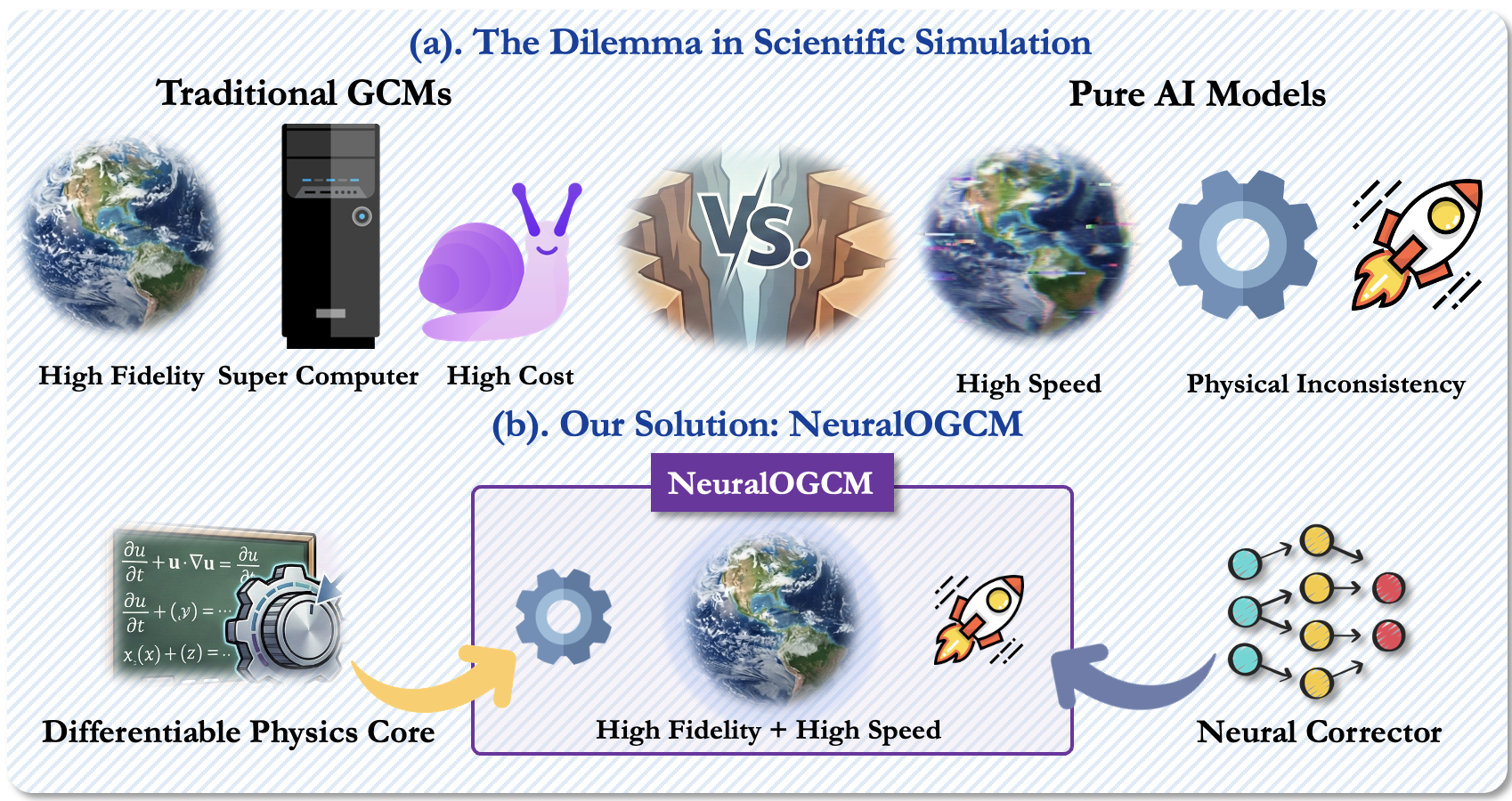}
\caption{
    The motivation for our model, \method{}. 
    \textbf{(a)} A dilemma exists between traditional GCMs, which are physically faithful but slow, and pure AI models, which are fast but prone to long-term instability.
    \textbf{(b)} \method{} bridges this gap by integrating a \textit{Differentiable Physics Core} with a \textit{Neural Corrector}. 
    Its key innovation is the learnable physics core, where physical parameters are optimized end-to-end from data, achieving both high fidelity and computational speed.
}
  \label{fig:intro}
\end{figure}
Climate change is one of the most severe global challenges of our time~\cite{lee2025observed, kochkov2024neural, wu2025advanced}. 
Within this complex system, the ocean~\cite{shi2024oceanvp, parasyris2025marine} plays a pivotal role as the Earth's primary climate regulator and buffer, where its dynamics profoundly influence global weather patterns, ecosystems, and human societies. 
Accurate and efficient simulation of ocean circulation is therefore crucial for advancing our scientific understanding and informing future decisions~\cite{madec1997ocean, parasyris2025marine}. 
Currently, Ocean General Circulation Models (OGCMs) serve as the core scientific tool for this purpose.

However, existing ocean modeling approaches face a fundamental dilemma: a sharp trade-off between computational efficiency and physical fidelity. 
On one hand, traditional GCMs~\cite{madec1997ocean, smith1992parallel, grotch1991use, wilby1997downscaling}, based on first principles of physics, represent the gold standard for long-term stability and physical reliability. 
Yet, their computational cost is prohibitively high, often requiring months of supercomputer time to simulate a single century-scale climate scenario. This severely limits the scope of scientific inquiry, such as performing large-scale ensemble forecasts or uncertainty quantification. 
On the other hand, pure data-driven models, particularly those based on deep learning, achieve orders-of-magnitude speedups~\cite{wu2024earthfarsser, wu2024neural, wu2024pastnet, wu2024prometheus, wu2024pure,gao2025oneforecast, li2022fourier}. 
However, their black-box nature, lack of physical consistency, and tendency to accumulate errors during long-term autoregressive rollouts leading to numerical divergence or physically nonsensical "hallucinations" make them difficult to trust for high-stakes scientific discovery~\cite{wu2025advanced}. 
This creates an enduring gap between computational speed and physical fidelity that hinders progress in climate science, a dilemma we illustrate in Figure~\ref{fig:intro}.

To realize this vision, we propose \method{}, which, \textit{\textbf{to the best of our knowledge, is the first hybrid ocean general circulation model to integrate differentiable physics programming with deep learning. }} Its concept is shown in Figure~\ref{fig:intro}. The core of \method{} consists of two synergistic components: a \textit{Differentiable Physics Core} that handles large-scale evolution based on fluid dynamics equations, with its key physical parameters being learnable (symbolized by the dial); and a \textit{Neural Corrector} that learns to amend for complex subgrid-scale processes and discretization errors not captured by the physics core. The tendencies produced by both components are integrated via a unified ODE solver~\cite{kochkov2024neural, wu2024prometheus}, enabling the entire model to be trained end-to-end to achieve both high fidelity and high efficiency.

Our main contributions are threefold:

\ding{182} We propose and implement a new hybrid modeling paradigm of \textit{learnable physics}, where key physical parameters can be optimized from data via gradient descent.

\ding{183} We design and build \method{}, the first end-to-end differentiable hybrid ocean general circulation model, successfully applying this paradigm to a complex real-world scientific problem.

\ding{184} We demonstrate through extensive experiments on a benchmark dataset that  \method{} achieves significant advantages in speed, accuracy, and long-term stability over existing methods.

\begin{figure*}
\centering
\includegraphics[width=1\linewidth]{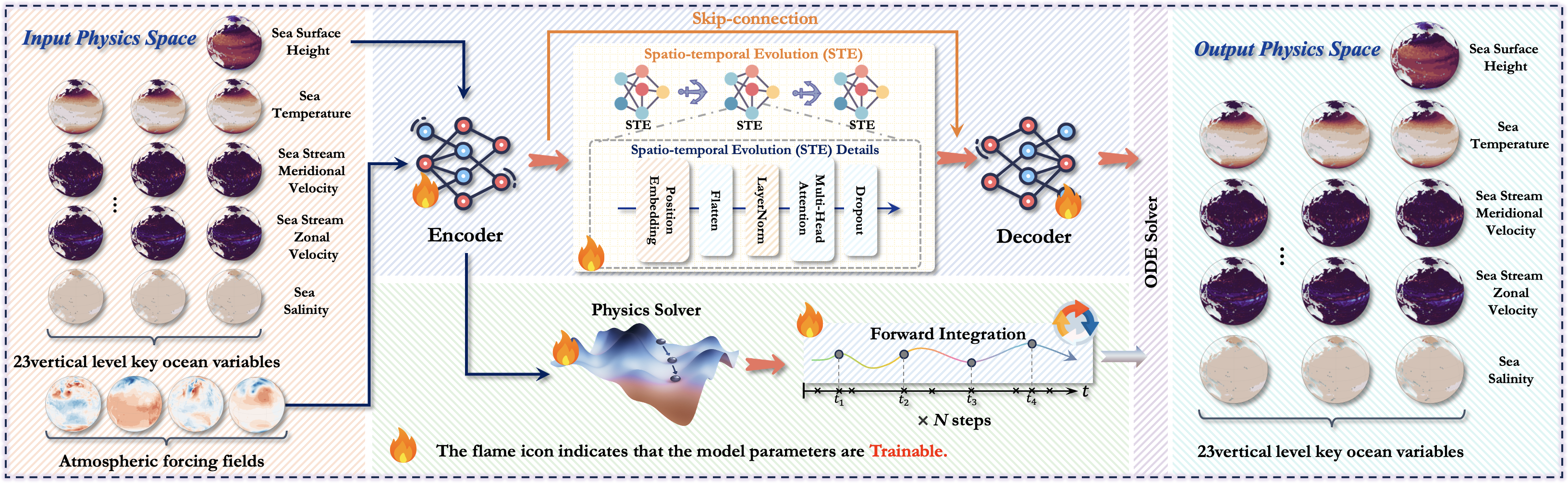}
\caption{
    Architecture of the Hybrid Physics-AI Ocean Model.
    The model predicts the future ocean state by combining tendencies from two synergistic branches. 
    A differentiable \textit{Physics Solver}, performing multi-step integration, computes the dynamic tendencies. 
    Concurrently, a data-driven branch, featuring an \textit{Encoder-Decoder} architecture and a \textit{Spatio-temporal Evolution (STE)} module, learns a corrective tendency. 
    Both tendencies are integrated by a final \textit{ODE Solver} to produce the output. 
    As indicated by flame icons, key physical parameters within the solver are learned end-to-end along with the neural network components.}
  \label{fig:NeuralOGCM_main}
\end{figure*}

\section{Methodology}

\subsection{Problem Definition}

Our goal is to learn an efficient operator $M$ for a spatio-temporal forecasting task, which maps the current state $(\mathbf{y}_t, \mathbf{F}_t)$ to a future state $\mathbf{y}_{t+\Delta t}$. Here, $\mathbf{y}_t \in \mathbb{R}^{C_o \times H \times W}$ represents the ocean's internal variables and $\mathbf{F}_t \in \mathbb{R}^{C_a \times H \times W}$ denotes the external atmospheric forcings. The system's evolution is governed by a set of partial differential equations (PDEs), which we model by decomposing the tendency operator $G$ into two synergistic components:
\begin{equation}\small
    \frac{\partial \mathbf{y}}{\partial t} = G(\mathbf{y}, \mathbf{F}) \approx \underbrace{\textcolor{PhysicsBlue}{G_{\text{phys}}(\mathbf{y}; \theta_p)}}_{\text{Differentiable Physics Core}} + \underbrace{\textcolor{NeuralOchre}{G_{\text{neural}}(\mathbf{y}, \mathbf{F}; \theta_n)}}_{\text{Neural Network Corrector}}
\end{equation}
The \textit{Differentiable Physics Core}, $G_{\text{phys}}$, approximates known dynamics such as advection and diffusion, but its key physical parameters, $\theta_p$, are learnable. The \textit{Neural Network Corrector}, $G_{\text{neural}}$, is a deep neural network parameterized by $\theta_n$ that learns the residual dynamics, including subgrid-scale effects and model errors. Our central challenge is to learn the total parameter set $\Theta = \{\theta_p, \theta_n\}$ end-to-end.

\subsection{The \method{} Framework}

Building upon the problem definition, we construct \textbf{\method}, as shown in Figure~\ref{fig:NeuralOGCM_main}, an end-to-end hybrid physics-AI framework. The core idea is to decompose the total tendency $G$ into a learnable physics-guided component and a data-driven one. The evolution of the state $\mathbf{y}$ over a time interval $\Delta t$ can be described by the integral form of its governing differential equation:
\begin{equation}\small
    \mathbf{y}_{t+\Delta t} = \mathbf{y}_t + \int_{t}^{t+\Delta t} G(\mathbf{y}(\tau), \mathbf{F}(\tau)) \, d\tau
    \label{eq:integral_form}
\end{equation}
In our framework, we approximate this integral using a single-step explicit Forward Euler scheme. This defines our model, \method, parameterized by $\Theta = \{\theta_p, \theta_n\}$, as the operator that performs this one-step integration:
\begin{equation}
\begin{split}
    \method(\mathbf{y}_t, \mathbf{F}_t; \Theta) & = \mathbf{y}_t + \Delta t \cdot G_{\text{phys}}(\mathbf{y}_t; \theta_p) \\
    & \quad + \Delta t \cdot G_{\text{neural}}(\mathbf{y}_t, \mathbf{F}_t; \theta_n)
\end{split}
\label{eq:ode_solver}
\end{equation}
This integration step seamlessly fuses the two branches. Crucially, since all components are fully differentiable, the entire model can be trained end-to-end by backpropagating the loss from the final prediction, which is the output of $\method$. The subsequent subsections detail the specific designs of $G_{\text{phys}}$ and $G_{\text{neural}}$.

\subsection{Differentiable Physics Core}

To embed physical knowledge within a deep learning framework, we construct a fully differentiable dynamical solver that approximates key processes from the ocean's primitive equations. Our physics core explicitly simulates three primary dynamical processes: \textit{Advection}, which describes how quantities are transported with the velocity field; the \textit{Coriolis Force}, which accounts for deflection effects due to the Earth's rotation; and \textit{Horizontal Diffusion}, which represents the effects of molecular viscosity and subgrid-scale turbulent mixing.

We implement these differential operators using standard operations in deep learning frameworks. For instance, spatial gradients $\nabla$ are approximated via 2D convolutions with finite difference kernels. To handle the spherical grid, we incorporate latitude-dependent metric factors in our calculations and apply periodic padding for boundaries in the longitudinal direction. Traditionally, the horizontal diffusion coefficient $\nu$ is a hyperparameter set manually by experts, which greatly influences the model's energy dissipation and long-term stability, making it a major source of uncertainty. In our framework, we define these coefficients as learnable parameters of the model, $\theta_p = \{\nu_{\text{momentum}}, \nu_{\text{tracer}}\}$. To ensure physical plausibility (diffusion must be non-negative), we apply a \textit{softplus} function to these parameters before their use in the solver. By exposing these physical parameters to the optimizer, \method{} can autonomously learn the effective diffusion from data, thereby partially correcting deficiencies in the subgrid-scale parameterization scheme.

\subsection{Neural Network Corrector}
The objective of the Neural Network Corrector, $G_{\text{neural}}$, is to learn the residual dynamics missed by the physics core. Its architecture is an Encoder-Decoder network featuring a Spatio-temporal Evolution (STE) module.

First, the Encoder maps the concatenated state $(\mathbf{y}_t, \mathbf{F}_t)$ to a low-dimensional latent state $\mathbf{z}_t$ through a series of convolutions and downsampling operations, while extracting multi-scale feature maps $\mathbf{s}_t$ for skip-connections.
\begin{equation}
    \mathbf{z}_t, \mathbf{s}_t = \text{Encoder}(\text{concat}(\mathbf{y}_t, \mathbf{F}_t); \theta_{\text{enc}})
\end{equation}
In the network's bottleneck, the STE module processes the latent state $\mathbf{z}_t$ using a multi-head self-attention mechanism to capture long-range spatial dependencies. The input feature map $\mathbf{z}_t \in \mathbb{R}^{C \times H \times W}$ is flattened into a sequence $X \in \mathbb{R}^{N \times d_{\text{model}}}$, where $N=H \times W$ and $d_{\text{model}}=C$.

The module computes $h$ attention heads in parallel. The query ($Q_i$), key ($K_i$), and value ($V_i$) for each head are obtained by linearly projecting the input sequence $X$:
\begin{equation}\small
    Q_i = X W_i^Q, \quad K_i = X W_i^K, \quad V_i = X W_i^V
\end{equation}
The output of each head is computed using scaled dot-product attention:
\begin{equation}\small
    \text{head}_i = \text{Attention}(Q_i, K_i, V_i) = \text{softmax}\left(\frac{Q_i K_i^T}{\sqrt{d_k}}\right)V_i
\end{equation}
The outputs of all heads are concatenated and passed through a final linear projection to produce the module's output:
\begin{equation}\small
    \text{MultiHead}(X) = \text{Concat}(\text{head}_1, \dots, \text{head}_h)W^O
\end{equation}
This process yields the evolved latent state $\mathbf{z}'_t$:
\begin{equation}
    \mathbf{z}'_t = \text{STE}(\mathbf{z}_t; \theta_{\text{ste}})
\end{equation}

Finally, the Decoder takes the evolved latent state $\mathbf{z}'_t$ and the skip-connection features $\mathbf{s}_t$ as input, reconstructing the correction tendency tensor $G_{\text{neural}}$ via a series of transposed convolutions and upsampling operations.
\begin{equation}
    G_{\text{neural}} = \text{Decoder}(\mathbf{z}'_t, \mathbf{s}_t; \theta_{\text{dec}})
\end{equation}
The set of trainable parameters for the neural network corrector is thus $\theta_n = \{\theta_{\text{enc}}, \theta_{\text{ste}}, \theta_{\text{dec}}\}$.

\subsection{Training Details}
The model is trained by minimizing the mean squared error (MSE) between the single-step prediction and the ground truth. The physics core operates in physical space, while the neural network and loss are computed in a normalized space. This requires denormalizing the input for the physics core and re-normalizing its output before combining it with the neural network's tendency during each forward pass. The loss function is defined as:
\begin{equation}\small
    L(\Theta) = \mathbb{E}_{(\mathbf{y}_t, \mathbf{F}_t), \mathbf{y}_{t+\Delta t} \sim \mathcal{D}} \left\| M(\mathbf{y}_t, \mathbf{F}_t; \Theta) - \mathbf{y}_{t+\Delta t} \right\|_2^2
\end{equation}
where $\mathcal{D}$ represents the training dataset, and the loss is computed in the normalized space. We use the AdamW optimizer for end-to-end training with a learning rate of 1e-4, a weight decay of 1e-5, and a batch size of 2.
\section{Experiments}

\subsection{Experimental Setup}

\noindent\textbf{Datasets.}
We evaluate our proposed \method{} model on a large-scale global ocean simulation task. The study utilizes a dataset that fuses two authoritative reanalysis sources: {GLORYS12} for oceanic state variables and {ERA5} for atmospheric forcing fields. For the computational requirements of our experiments, all variables are resampled to a {1.5°} spatial resolution, while maintaining a 24-hour (daily mean) temporal resolution. Our model is designed to predict five key oceanic state variables: sea temperature ($T_o$), salinity ($S$), zonal velocity ($U_o$), and meridional velocity ($V_o$) across 23 vertical levels, along with sea surface height (SSH) at a single level. The model is driven by four atmospheric variables. Furthermore, a static land-sea mask (LSM) is employed to define the ocean domain. Specific details of the dataset variables are summarized in Table~\ref{tab:dataset}.

\medskip
\noindent\textbf{Data Partitioning and Implementation Details.}
The entire dataset spans the years 1993 to 2020. We follow a standard chronological split for time-series forecasting, using data from {1993--2017} for training, {2018--2019} for validation, and {2020} for testing.

Our model is implemented using the PyTorch framework, and all experiments are conducted on {8 NVIDIA A100 GPUs} using a data parallelism strategy. For training, we use the AdamW optimizer with a learning rate of 1e-4 and a weight decay of 1e-5. The batch size is set to 2 per GPU, resulting in an effective total batch size of 16.


\begin{table}[h!] 
\caption{Details of the dataset for the global ocean simulation task.}
\label{tab:dataset}
\centering 
\small 
\begin{tabular}{llccc}
\toprule
\textbf{Type} & \textbf{Abbr.} & \textbf{Layers} & \textbf{Time Res.} & \textbf{Spatial Res.} \\
\midrule
Atmospheric & U10M & 1 & 24h & 1.5° \\
Atmospheric & V10M & 1 & 24h & 1.5° \\
Atmospheric & T2M & 1 & 24h & 1.5° \\
Atmospheric & MSLP & 1 & 24h & 1.5° \\
\midrule
Oceanic & S & 23 & 24h & 1.5° \\
Oceanic & $U_o$ & 23 & 24h & 1.5° \\
Oceanic & $V_o$ & 23 & 24h & 1.5° \\
Oceanic & $T_o$ & 23 & 24h & 1.5° \\
Oceanic & SSH & 1 & 24h & 1.5° \\
\midrule
Static & LSM & --- & --- & 1.5° \\
\bottomrule
\end{tabular}
\end{table}

\subsection{Baseline Models}

We test our \method{} against a series of advanced, purely data-driven spatio-temporal forecasting models. These baselines include the classic {U-Net} and {ConvLSTM}, the efficient general-purpose model {SimVP}, and leading large-scale weather forecasting models: {FourCastNet}. The comparison focuses on evaluating performance in terms of simulation accuracy and long-term autoregressive stability.

\subsection{Main Results}

We evaluate the performance of \method{} against state-of-the-art baselines on the global ocean simulation task. Table~\ref{tab:rmse_styled} presents the quantitative comparison using Root Mean Squared Error (RMSE) across different lead times ranging from 10 to 120 days.

As shown in the table, \method{} consistently outperforms all baseline models. In short-term forecasting (10 days), our model achieves the lowest RMSE of $0.919$, demonstrating superior accuracy. The performance gap becomes increasingly pronounced in long-term simulations. At a lead time of 120 days, \method{} maintains a low RMSE of $1.574$, significantly surpassing the second-best model, U-Net ($1.886$). Notably, while large-scale pure AI models like FourCastNet suffer from severe error accumulation and instability in long-term rollouts ($3.332$ at 120 days), \method{} remains stable and physically consistent. These results validate that integrating a differentiable physics core effectively mitigates the long-term divergence issues common in purely data-driven approaches.

\begin{figure}[h!]
  \centering
  \includegraphics[width=0.95\linewidth]{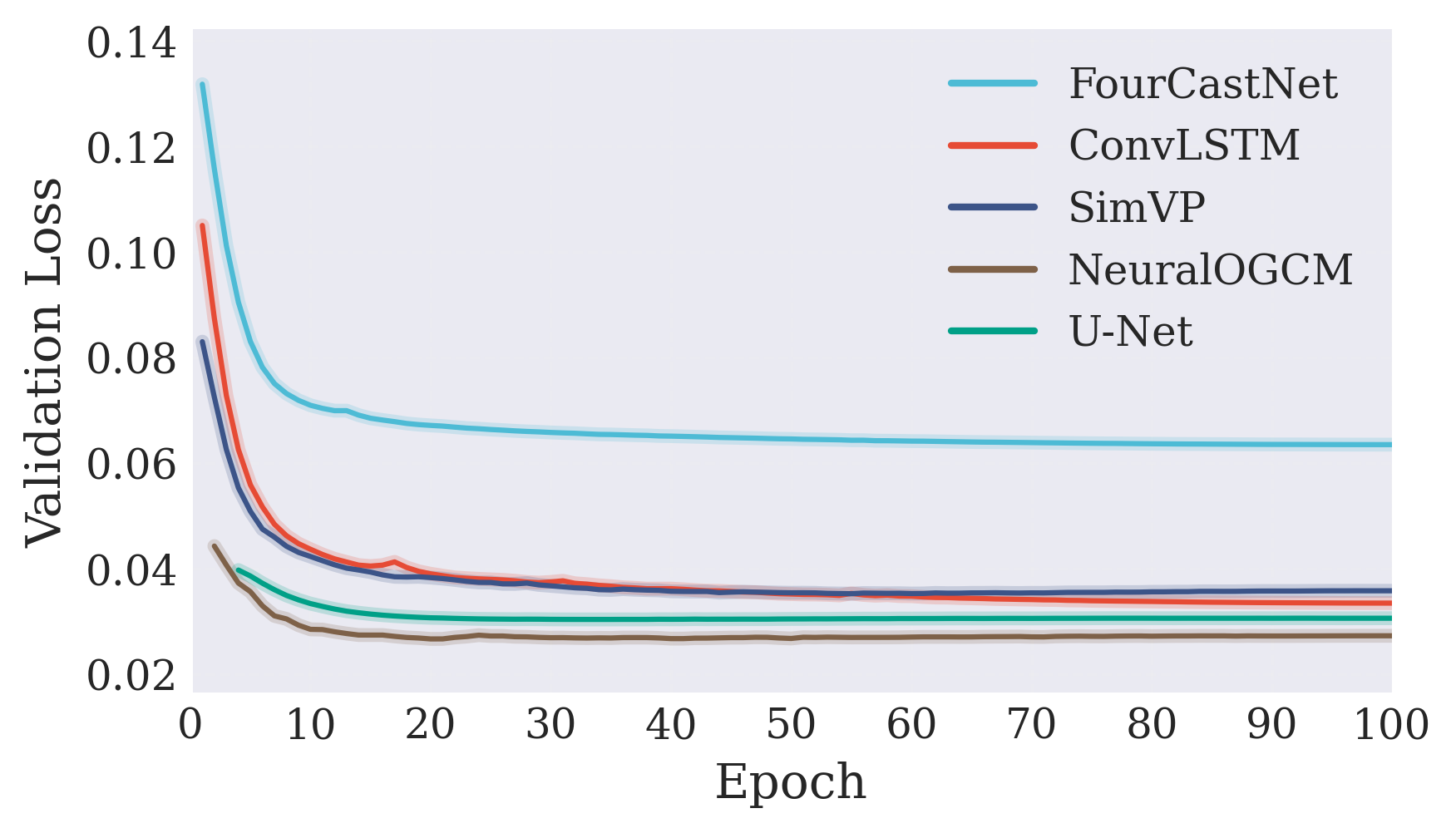}
\caption{Validation loss curves of \method{} compared with baseline models during the training process, demonstrating faster convergence and lower final error.}
  \label{fig:loss}
\end{figure}

Figure~\ref{fig:loss} further illustrates the convergence behavior during training, where \method{} achieves lower validation loss compared to other methods.


\begin{table}[t!]
\caption{
    Comparison of autoregressive simulation performance using Root Mean Squared Error (RMSE). Lower values are better. The best-performing model is highlighted.
}
\label{tab:rmse_styled}
\small
\centering
\definecolor{highlightgray}{gray}{0.9}
\definecolor{stdcolor}{gray}{0.5}

\setlength{\tabcolsep}{3pt}

\sisetup{
  table-format=1.3,
  detect-weight=true,
  detect-inline-weight=math
}

\resizebox{\linewidth}{!}{
    \begin{tabular}{
        l
        S S S S
    }
    \toprule
    \multirow{2}{*}{\textbf{Methods}} & \multicolumn{4}{c}{\textbf{Lead Time (days)}} \\
    \cmidrule(lr){2-5}
    & {10} & {30} & {60} & {120} \\
    \midrule
    U-Net       & {$0.923_{\textcolor{stdcolor}{0.012}}$} & {$1.436_{\textcolor{stdcolor}{0.018}}$} & {$1.719_{\textcolor{stdcolor}{0.026}}$} & {$1.886_{\textcolor{stdcolor}{0.024}}$} \\
    ConvLSTM    & {$0.979_{\textcolor{stdcolor}{0.017}}$} & {$1.565_{\textcolor{stdcolor}{0.030}}$} & {$2.053_{\textcolor{stdcolor}{0.036}}$} & {$2.840_{\textcolor{stdcolor}{0.046}}$} \\
    SimVP       & {$0.951_{\textcolor{stdcolor}{0.016}}$} & {$1.414_{\textcolor{stdcolor}{0.026}}$} & {$1.705_{\textcolor{stdcolor}{0.028}}$} & {$1.946_{\textcolor{stdcolor}{0.021}}$} \\
    FourCastNet & {$0.959_{\textcolor{stdcolor}{0.018}}$} & {$1.424_{\textcolor{stdcolor}{0.030}}$} & {$2.433_{\textcolor{stdcolor}{0.141}}$} & {$3.332_{\textcolor{stdcolor}{0.066}}$} \\
    \midrule
    \rowcolor{highlightgray}
    \textbf{\method{}} & {\textbf{$0.919_{\textcolor{stdcolor}{0.015}}$}} & {\textbf{$1.280_{\textcolor{stdcolor}{0.020}}$}} & {\textbf{$1.439_{\textcolor{stdcolor}{0.027}}$}} & {\textbf{$1.574_{\textcolor{stdcolor}{0.037}}$}} \\
    \bottomrule
    \end{tabular}
}
\end{table}

\begin{figure}[h!]
  \centering
  \includegraphics[width=0.95\linewidth]{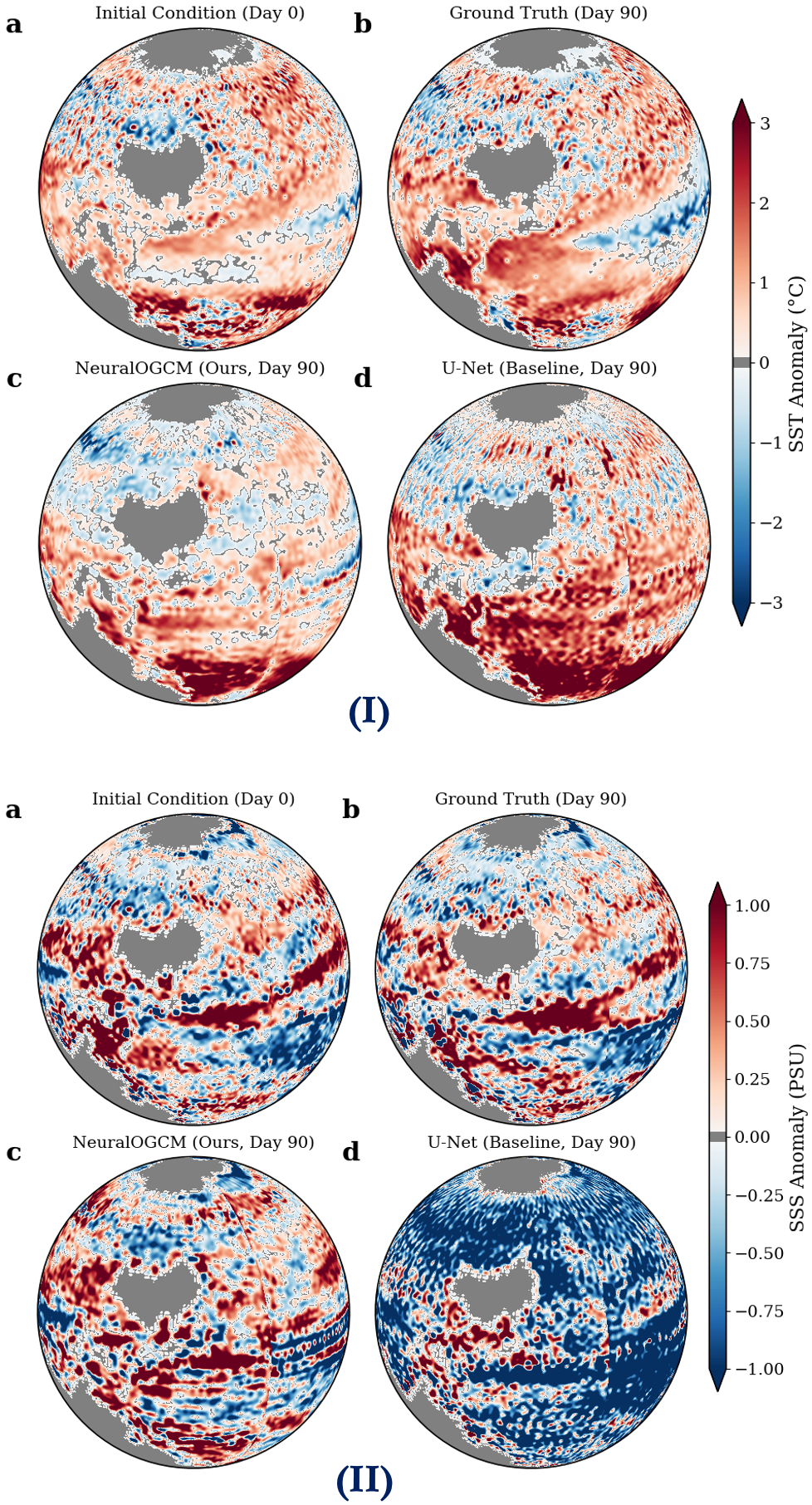} 
  \caption{
    Visualization of 90-day autoregressive rollout results for (I) Sea Surface Temperature (SST) Anomaly and (II) Sea Surface Salinity (SSS) Anomaly. 
    Columns show: (a) Initial condition at Day 0, (b) Ground Truth at Day 90, (c) Prediction by \method{}, and (d) Prediction by the U-Net baseline. 
    While \method{} maintains high physical fidelity and structural coherence consistent with the ground truth, the pure AI baseline exhibits severe error accumulation, manifested as over-saturated values in SST and non-physical high-frequency artifacts in SSS.
  }
  \label{fig:rollout_vis}
\end{figure}

\subsection{Long-term Autoregressive Rollout Analysis}

To intuitively evaluate the model's capability in maintaining physical consistency over extended horizons, we conduct a 90-day autoregressive rollout experiment. Figure~\ref{fig:rollout_vis} visualizes the spatial distribution of Sea Surface Temperature (SST) and Sea Surface Salinity (SSS) anomalies at the 90th simulation day.

As observed in panel (c), \method{} successfully preserves the complex spatial structures and accurate anomaly magnitudes compared to the Ground Truth (panel b). It accurately captures large-scale circulation patterns without exhibiting significant dissipation or non-physical drift. In sharp contrast, the purely data-driven baseline (U-Net, panel d) suffers from severe numerical instability. For SST (I), the baseline tends to overestimate extreme values, leading to saturated anomalies. More critically, in the SSS field (II), the baseline exhibits substantial high-frequency noise and grid-like artifacts, indicating a breakdown in physical realism due to error accumulation. These visual results corroborate that embedding the differentiable physics core effectively acts as a physical inductive bias, regularizing the long-term evolution and preventing the "hallucinations" common in pure deep learning approaches.

\section{Conclusions}
In this work, we introduce \method{}, the first differentiable ocean model, to our knowledge, that seamlessly integrates a learnable physics core with a neural network corrector in an end-to-end framework. This pioneering approach directly addresses the long-standing trade-off between computational speed and physical fidelity. Our experiments demonstrate that \method{} maintains long-term physical stability, significantly outperforms purely data-driven baselines in long-range accuracy, and operates orders of magnitude faster than traditional numerical models. Our work paves a new path for building fast, stable, and physically plausible scientific models, showcasing the immense potential of fusing differentiable programming with deep learning to advance complex Earth system modeling.

\bibliography{example_paper}

@String { ARXIV        = {arXiv} }

@String { AAAI         = {AAAI}}

@inproceedings{wu2024pastnet,
  title={Pastnet: Introducing physical inductive biases for spatio-temporal video prediction},
  author={Wu, Hao and Xu, Fan and Chen, Chong and Hua, Xian-Sheng and Luo, Xiao and Wang, Haixin},
  booktitle={Proceedings of the 32nd ACM international conference on multimedia},
  pages={2917--2926},
  year={2024}
}

@article{wu2025advanced,
  title={Advanced long-term earth system forecasting by learning the small-scale nature},
  author={Wu, Hao and Gao, Yuan and Shu, Ruiqi and Wang, Kun and Gou, Ruijian and Wu, Chuhan and Liu, Xinliang and He, Juncai and Cao, Shuhao and Fang, Junfeng and others},
  journal={arXiv preprint arXiv:2505.19432},
  year={2025}
}

@article{gao2025oneforecast,
  title={OneForecast: A Universal Framework for Global and Regional Weather Forecasting},
  author={Gao, Yuan and Wu, Hao and Shu, Ruiqi and Dong, Huanshuo and Xu, Fan and Chen, Rui and Yan, Yibo and Wen, Qingsong and Hu, Xuming and Wang, Kun and others},
  journal={arXiv preprint arXiv:2502.00338},
  year={2025}
}

@inproceedings{wu2024prometheus,
  title={Prometheus: Out-of-distribution Fluid Dynamics Modeling with Disentangled Graph ODE},
  author={Wu, Hao and Wang, Huiyuan and Wang, Kun and Wang, Weiyan and Ye, Changan and Tao, Yangyu and Chen, Chong and Hua, Xian-Sheng and Luo, Xiao},
  booktitle={Proceedings of the 41st International Conference on Machine Learning},
  pages={PMLR 235},
  year={2024},
  organization={PMLR},
  address={Vienna, Austria}
}

@article{wu2024pure,
  title={Pure: Prompt evolution with graph ode for out-of-distribution fluid dynamics modeling},
  author={Wu, Hao and Wang, Changhu and Xu, Fan and Xue, Jinbao and Chen, Chong and Hua, Xian-Sheng and Luo, Xiao},
  journal={Advances in Neural Information Processing Systems},
  volume={37},
  pages={104965--104994},
  year={2024}
}

@article{hersbach2020era5,
  title={The ERA5 global reanalysis},
  author={Hersbach, Hans and Bell, Bill and Berrisford, Paul and Hirahara, Shoji and Hor{\'a}nyi, Andr{\'a}s and Mu{\~n}oz-Sabater, Joaqu{\'\i}n and Nicolas, Julien and Peubey, Carole and Radu, Raluca and Schepers, Dinand and others},
  journal={Quarterly Journal of the Royal Meteorological Society},
  volume={146},
  number={730},
  pages={1999--2049},
  year={2020},
  publisher={Wiley Online Library}
}

@article{li2022fourier,
  title={Fourier neural operator with learned deformations for pdes on general geometries},
  author={Li, Zongyi and Huang, Daniel Zhengyu and Liu, Burigede and Anandkumar, Anima},
  journal={arXiv preprint arXiv:2207.05209},
  year={2022}
}

@inproceedings{wu2024earthfarsser,
  title={Earthfarsser: Versatile Spatio-Temporal Dynamical Systems Modeling in One Model},
  author={Wu, Hao and Liang, Yuxuan and Xiong, Wei and Zhou, Zhengyang and Huang, Wei and Wang, Shilong and Wang, Kun},
  booktitle={Proceedings of the AAAI Conference on Artificial Intelligence},
  volume={38},
  number={14},
  pages={15906--15914},
  year={2024}
}

@article{lee2025observed,
  title={Observed multi-decadal increase in the surface ocean’s thermal inertia},
  author={Lee, Chaehyeong and Song, Hajoon and Choi, Yeonju and Cho, Ajin and Marshall, John},
  journal={Nature Climate Change},
  pages={1--7},
  year={2025},
  publisher={Nature Publishing Group UK London}
}

@inproceedings{wu2024neural,
  title={Neural Manifold Operators for Learning the Evolution of Physical Dynamics},
  author={Wu, Hao and Weng, Kangyu and Zhou, Shuyi and Huang, Xiaomeng and Xiong, Wei},
  booktitle={Proceedings of the 30th ACM SIGKDD Conference on Knowledge Discovery and Data Mining},
  pages={3356--3366},
  year={2024}
}

@article{parasyris2025marine,
  title={Marine Heatwaves in the Mediterranean Sea: A Convolutional Neural Network study for extreme event prediction},
  author={Parasyris, Antonios and Metheniti, Vassiliki and Kampanis, Nikolaos and Darmaraki, Sofia},
  journal={Ocean Science},
  volume={21},
  number={3},
  pages={897--912},
  year={2025},
  publisher={Copernicus Publications G{\"o}ttingen, Germany}
}

@article{shi2024oceanvp,
  title={OceanVP: A HYCOM based benchmark dataset and a relational spatiotemporal predictive network for oceanic variable prediction},
  author={Shi, Zhensheng and Zheng, Haiyong and Dong, Junyu},
  journal={Ocean Engineering},
  volume={304},
  pages={117748},
  year={2024},
  publisher={Elsevier}
}

@article{kochkov2024neural,
  title={Neural general circulation models for weather and climate},
  author={Kochkov, Dmitrii and Yuval, Janni and Langmore, Ian and Norgaard, Peter and Smith, Jamie and Mooers, Griffin and Kl{\"o}wer, Milan and Lottes, James and Rasp, Stephan and D{\"u}ben, Peter and others},
  journal={Nature},
  volume={632},
  number={8027},
  pages={1060--1066},
  year={2024},
  publisher={Nature Publishing Group UK London}
}

@article{madec1997ocean,
  title={Ocean general circulation model reference manual},
  author={Madec, Gurvan and Delecluse, Pascale and Imbard, M and Levy, C},
  journal={Note du P{\^o}le de mod{\'e}lisation},
  volume={495},
  year={1997}
}

@article{smith1992parallel,
  title={Parallel ocean general circulation modeling},
  author={Smith, RD and Dukowicz, JK and Malone, RC},
  journal={Physica D: Nonlinear Phenomena},
  volume={60},
  number={1-4},
  pages={38--61},
  year={1992},
  publisher={Elsevier}
}

@article{grotch1991use,
  title={The use of general circulation models to predict regional climatic change},
  author={Grotch, Stanley L and MacCracken, Michael C},
  journal={Journal of climate},
  pages={286--303},
  year={1991},
  publisher={JSTOR}
}

@article{wilby1997downscaling,
  title={Downscaling general circulation model output: a review of methods and limitations},
  author={Wilby, Robert L and Wigley, Thomas ML},
  journal={Progress in physical geography},
  volume={21},
  number={4},
  pages={530--548},
  year={1997},
  publisher={Sage Publications Sage CA: Thousand Oaks, CA}
}
\bibliographystyle{icml2026}



\newpage
\appendix
\onecolumn
\section{Dataset Details}
\subsection{Dataset}

In the large-scale global ocean simulation task, the atmosphere variables are sourced from  ERA5~\cite{hersbach2020era5} dataset and the ocean variables are sourced from the GLOYRS12 dataset. ERA5 offers global atmosphere state, and the selected subset contains 4 variables (U10M, V10M, T2M, MSLP) with surface level, which can be downloaded from \url{https://cds.climate.copernicus.eu}. GlORYS12 offers daily mean data covering latitudes between -80° and 90° from 1993 to the present, and the subset we use includes 4 depth level ocean variables (each with 23 depth levels, corresponding to 0.49 m, 2.65 m, 5.08 m, 7.93 m, 11.41 m, 15.81 m, 21.60 m, 29.44 m, 40.34 m, 55.76 m, 77.85 m, 92.32 m, 109.73 m, 130.67 m, 155.85 m, 186.13 m, 222.48 m, 266.04 m, 318.13 m, 380.21 m, 453.94 m, 541.09 m and 643.57 m), Sea salinity (S), Sea stream zonal velocity ($\mathrm{U}_{\mathrm{o}}$), Sea stream meridional velocity ($\mathrm{V}_{\mathrm{o}}$), Sea temperature ($\mathrm{T}_{\mathrm{o}}$), and 1 surface level variable Sea surface height (SSH), which can be downloaded from \url{https://data.marine.copernicus.eu}. For data partitioning, we use years from 1993 to 2020, which are 1993-2017 for training, 2018-2019 for validating, and 2020 for testing. To improve computational efficiency, we use bilinear interpolation to downsample them to 1.5 degree (H=121, W=240) spatial resolution. And to better adapt to the input of different architecture models, we use the data with size 120 × 240. The temporal resolution is 24 hours, which corresponds to 12:00 UTC for atmosphere variables and daily mean state for ocean variables.

\subsection{Data Preprocessing}
Different atmosphere and ocean variables exhibit substantial variations in magnitude. To enable the model to concentrate on accurate simulation rather than learning the inherent magnitude discrepancies among variables, we normalize the input data prior to model ingestion. Specifically, for atmosphere variables, we calculate these statistics from an extended dataset covering the period 1993 to 2017. For ocean variables, we first compute the climatological mean of all periodic variables using data from 1993–2017 (the first 365 days of each year in the training set). The shape of the climatological mean is (365, 47, 121, 240). Specifically, 365 denotes the days, 47 denotes the number of periodic variables (ie., 23 layers Sea salinity, 23 layers Sea temperature, and Sea surface height). 121 denotes the height and 240 denotes the width. Based on this mean, Sea salinity, Sea temperature, and Sea surface height are converted into Sea salinity anomaly, Sea temperature anomaly, and Sea surface height anomaly, respectively. We then compute the mean and standard deviation of all variables (using the anomaly fields for periodic variables) over the same 1993–2017 training period, and use these statistics for subsequent normalization. Each variable thus possesses a dedicated mean and standard deviation. Before inputting data into the model, we normalize the data by subtracting the corresponding mean and dividing the respective standard deviation. For the `nan' values of land, we fill them with zero before inputting the data into the model.

\end{document}